# An anthropomorphic continuum robotic neck actuated by SMA spring-based multipennate muscle architecture

Ratnangshu Das, Yashaswi Sinha, Anirudha Bhattacharjee, and Bishakh Bhattacharya

*Abstract*— This work presents a novel Shape Memory Alloy spring actuated continuum robotic neck that derives inspiration from pennate muscle architecture. The proposed design has 2DOF, and experimental studies reveal that the designed joint can replicate the human head's anthropomorphic range of motion. We enumerate the analytical modelling for SMA actuators and the kinematic model of the proposed design configuration. A series of experiments were conducted to assess the performance of the anthropomorphic neck by measuring the range of motion with varying input currents. Furthermore, the experiments were conducted to validate the analytical model of the SMA Multiphysics and the continuum backbone. The existing humanoid necks have been powered by conventional actuators that have relatively low energy efficiency and are prone to wear. The current research envisages application of non-conventional actuator such as SMA springs with specific geometric configuration yielding high power to weight ratio that delivers smooth motion for continuum robots as demonstrated in this present work.

## I. INTRODUCTION

Humanoid robots have become popular in the last two decades and several mechanical designs has been proposed to replicate the anthropomorphic behavior of the human neck [1,3,4,5]. The human head exhibits a complex spatial motion due to the cervical portion of the spine working in conjunction with the muscles. There are seven cervical vertebrae of the spine in the neck, and each has 6DOF [2]. To synthesize mechanisms with reduced complexity only 3DOFs of the neck behavior viz. roll, pitch and yaw are generally considered. Based on the configurations the humanoid neck mechanisms can be broadly classified into two categories serial and parallel. In serial configuration, each DOF consist of a rigid link and is independently actuated whereas in parallel, the moving platform and the base is connected by several tendons or flexible elements that are usually independently actuated. In recent years several works have also been proposed which harnesses the strain energy of the deformed structural elements to produce a desired output motion. These encompasses a broad range of robots termed as soft robots and are synthesized with adaptive compliance. Some examples of robots with motor actuated 2DOF (yaw and pitch movements only) neck are NAO [4], Pepper[5], ASIMO[1], HUBO[3], and that with 3DOF(roll, pitch and yaw) are HRP-4C [6] and iCub[7].

Ratnangshu Das, Yashaswi Sinha and Anirudha Bhattcharjee are with Department of Mechanical Engineering, Indian Institute of Technology Kanpur, (e-mail: rdas@iitk.ac.in; ysinha@iitk.ac.in; anirub@iitk.ac.in ).

Dr. Bishakh Bhattacharya is with Department of Mechanical Engineering, Indian Institute of Technology Kanpur, (phone: (+91) 512 259 7913; e-mail: bishakh@iitk.ac.in ).

The existing humanoid necks have been powered by conventional actuators [8]. Not only the conventional actuators are prone to wear and are expensive, but they also have a low power to weight ratio. To overcome the limitations of conventional actuators, the current work enumerates a novel bioinspired Shape Memory Alloy (SMA) spring based anthropomorphic robot neck mechanism. Smart Materials are special class of materials known to have stimuli other than external stress to obtain strain. SMA belongs to this special class of materials, where the material regains its shape when subjected to high temperature. The other differentiating factor of SMAs are their high actuation energy density, when compared to other similar materials where the coupling is direct. Furthermore, SMAs exhibit hysteresis behavior when administered under cyclical loading, where it can both absorb and dissipate mechanical energy. The distinctive characteristics of Shape Memory Alloys makes them ideal for both actuation and sensing applications.

Mammalian musculature have been the source of inspiration for a multitude of bio-inspired actuators. There are a variety of muscle architecture present in the mammalian body, with each architecture assigned a specific role. Though, generally the entire muscles can be assigned into two major sections: parallel and pennate. While in pennate musculature, the muscle fibers are aligned at an oblique angle to the tendon, in the parallel musculature the muscles are aligned parallel to each other. As a consequence of this the maximum strain is observed in the parallel muscle architecture, as the entire strain of the muscle fiber axis aligned. Meanwhile, in the case of pennate muscle architecture, since the muscle fibers are not aligned to the central tendon, the entire fiber strain is not observed along the tendon. However, due to this architecture more fiber can be fit into a given area and hence a higher force can be generated. Therefore, pennate architecture is preferred over a parallel architecture where a higher force is required in a given cross sectional area. Furthermore, pennate muscles have tendons that may run down the middle (bipennate), along one side (unipennate) or multiple tendons converging towards a central junction (multipennate). Additionally, the arrangement of the SMA wires have impact on the force and the stroke obtained on actuating. The current research describes a series of SMA actuator arranged in an orientation (where the central structure of each actuator is converging at a certain junction) at of fibers present in a multipennate muscle architecture, thus giving a higher flexibility and improving the performance.

In this work, we propose a design of SMA spring actuated robotic neck mechanism. The springs are arranged in a multi-pennate muscle architecture thereby providing enhanced payload capacity. The design incorporates a fixed base and a mobile platform which holds the head. The pennate arrangements of the actuators enable the head to achieve 2DOF anthropomorphic range of motions of the human head as illustrated in Fig. 1.

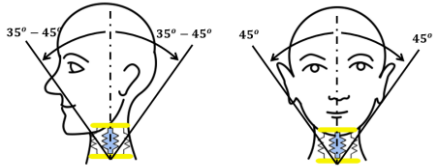

Fig.1. The human neck range of motion (a) forward flexion and backward extension (tilt or pitch) (b) lateral bending (roll) [9]

The paper has been arranged into the following sections II Methodology, describes the proposed design and corresponding analytical models. Section III Results, discusses the analytical results and the experimental validations. IV Discussions, highlights the unique advantages of the use of the proposed bio-inspired, non-conventional actuated mechanism and V Conclusions, discuss the limitations and future scope of the development.

## II. METHODOLOGY

### A. Design and Prototyping

As a proof of concept, an SMA-based multipennate neck has been developed and validated against the set of constitutive equations. Three different materials have been used in the proof of concept, each serving a particular purpose. Poly-Carbonate, having high strength and glass transition temperature, makes it suitable for working making the head mount and base plate, as shown in Fig 2(a). Furthermore, the muscle fibers have been emulated using SMA springs and *Flexinol®* from *Dynalloy, Inc*. Additionally, since the backbone needs low flexural stiffness, Thermoplastic Polyurethane (TPU) material has been used to manufacture the flexible backbone. It is imperative to note that the head mount, base plate, and flexible spine have all been built using additive manufacturing.

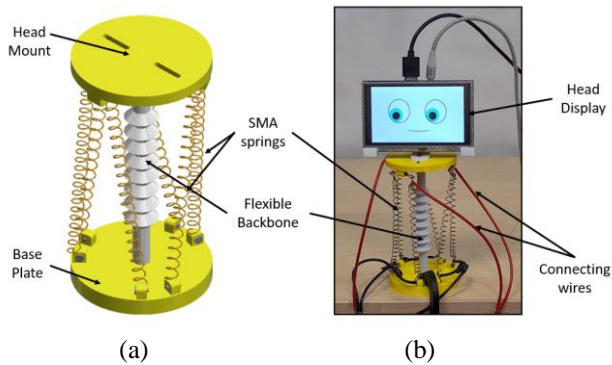

Fig 2: The proposed Shape-Memory-Alloy-based anthropomorphic neck has a multipennate configuration, where the wires are arranged obliquely for each pennate unit. (a) The head display depicts the CAD model of the multipennate neck (b) Front view of the multipennate neck.

### B. Analytical Modelling

The shape memory behavior depends on thermo-mechanical properties, which change as per the changes in macroscopic crystal structures associated with the phase transformation. Additionally, since the structure consists of a continuum backbone and SMA springs arranged in multipennate architecture, the analytical modeling requires the multipennate stiffness equation and continuum backbone kinematics on top of the SMA modeling.

#### 1) Constitutive Equation

The constitutive model develops a relation between stress rate ($\dot{\sigma}$), strain rate ($\dot{\epsilon}$) and martensite volume fraction rate ($\dot{\xi}$) along with the effect of the rate of temperature change ($\dot{T}$) on the stress rate. The constitutive model was birthed by Tanaka [10] and went on to be adopted by Brinson [11].

$$\left. \begin{array}{c} \dot{\tau} = A\dot{\gamma} + B\dot{\xi} + C\dot{T} \\ A = \dfrac{d^4 G}{8nD^3}; B = \dfrac{\pi d^3 \Omega}{8\sqrt{3}D}; C = \dfrac{\pi d^3 \theta_T}{8\sqrt{3}D} \end{array} \right\} \quad (1)$$

$$E = E_m \cdot \xi + E_a \cdot (1-\xi); G = \frac{E}{2(1+\nu)} \quad (2)$$

Where, $\theta_T$ is the thermal expansion factor, and $\Omega$ accounts for the effect of phase transformation change on the stress rate. $E$ is the phase-dependent Young's modulus of SMA, where $E_m$ and $E_a$ are Young's modulus of SMA in Martensite and Austenite phases, respectively.

#### 2) Phase Transformation Equation

The phase transformation function has been modeled by numerous researchers, with the general distinguishing factor being the choice of internal state variable and the evolution equation [12]. Different transformation models have been established, such as the Smooth transformation hardening model, the Cosine model, and the Exponential model (Tanaka). The cosine function employed in this phase transformation model is the same as that used by Liang and Rodgers and was developed by Elahinia and Ahmadian [13,14]. The model effectively captures the Shape Memory behavior under complex thermomechanical loading.

##### a) Reverse Transformation (Martensite to Austenite)

The equation governing the reverse transformation (Low-temperature phase to High-temperature phase) is:

$$\xi = 0; T < A_s' \quad (3)$$
$$\xi = \frac{\xi_m}{2}[\cos[a_a(T - A_s) + b_a \sigma] + 1]; A_s' < T < A_f'$$
$$\xi = 1; T > A_f'$$

Where, $a_a = \frac{\pi}{A_f - A_s}$; $b_a = -\frac{a_a}{C_a}$; $C_a$ is curve fitting parameter, $A_s' = A_s + \frac{\sigma}{C_a}$, $A_f' = A_f + \frac{\sigma}{C_a}$ and $\xi_m$ is the martensite volume fraction before the transformation phase.

##### b) Forward Transformation (Austenite to Martensite)

The equation governing the forward transformation (High temperature phase to Low temperature phase) is:

$$\xi = 0; T < M_f'$$
$$\xi = \frac{1-\xi_a}{2}\left[\cos[a_m(T - M_f) + b_m\sigma] + \frac{1+\xi_a}{2}\right]; M_f' < T < M_s' \quad (4)$$
$$\xi = 1; T > M_s'$$

Where, $a_m = \frac{\pi}{M_s - M_f}$; $b_m = -\frac{a_m}{C_m}$; $C_{am}$ is curve fitting parameter, $M_f' = M_f + \frac{\sigma}{C_m}$, $M_s' = M_s + \frac{\sigma}{C_m}$ and $\xi_a$ is the austenite volume fraction before the transformation phase.

*3) Energy Balance Equation:*
An input current signal provides the energy required to elevate the temperature of the springs, $I_{in}$, through the Joule heating phenomenon. The heat energy balance equation is given by:

$$m_{sp}C_p\dot{T} = I_{in}^2 R - A_c h_T(T - T_\infty) + m_{sp}\Delta H\dot{\xi} \quad (5)$$

Where, $R = (R_m\xi + R_a(1 - \xi))$ is the phase-dependent resistance, $\Delta H$ is the latent heat of transformation and $h_T$ is the convective heat loss coefficient.

*4) Continuum Backbone Kinematics*
The continuum backbone has been approximated as an arc with a constant curvature [15]. This allows us to parameterize the backbone using the arc parameters $[\kappa, \phi]$, where $\kappa$ is the curvature and $\phi$ is the angle that the bending plane makes with the x-axis, as illustrated in fig 2.

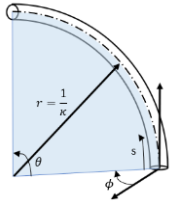

Fig 3: Kinematic framework using arc parameters $\kappa$ and $\phi$

The position vector of each arc point as a function of the distance along the arc, s, can be found geometrically and is represented by:

$$p(s) = \begin{bmatrix} \frac{\cos(\phi)}{\kappa}(1 - \cos(\kappa s)) \\ \frac{\sin(\phi)}{\kappa}(1 - \cos(\kappa s)) \\ \frac{1}{\kappa}\sin(\kappa s) \end{bmatrix} \quad (6)$$

The resultant transformation matrix finally turns out to be:

$$T(s) = \begin{bmatrix} R_z(\phi) \cdot R_y(\kappa s) \cdot R_z(\epsilon - \phi) & p(s) \\ 0 & 1 \end{bmatrix} \quad (7)$$

Where $\epsilon$ is the twist angle when the backbone is subjected to torsion.
Note that $\theta = \kappa l$, where l is the length of the backbone.

*5) Multipennate Stiffness Equation*
The SMA springs are arranged in a multipennate architecture, which can be imagined as a set of three muscles structures positioned at $120^o$ from each other. Each muscle structure comprises two SMA springs, which are aligned in a bipennate configuration.

Each muscle structure applies an actuation force of $F_k$ (k = 1, 2, 3) and the net moment $M_k$ produced by this force on the backbone can be calculated as:

$$M_k = (p_k(l) - p(l)) \times F_k \quad (11)$$

where $p_k(l)$ is the position vector of the point where the k$^{th}$ muscle applies force on the head mount and $p(l)$ is the position vector where the central backbone intersects with the head mount.

The force $F_k$, due to the pennate structure can be calculated from the force generated by each SMA spring, $F_f$ and the angle of pennation, $\alpha$, as:

$$F_k = 2F_f \cos(\alpha) \quad (12)$$

The multipennate architecture is illustrated in Fig. 3.

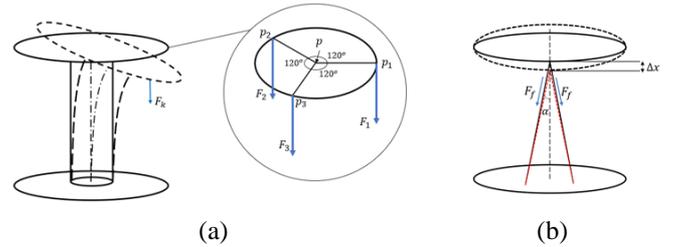

(a) (b)
Fig 4: Schematic of the SMA-based anthropomorphic neck. (a) The point of application and the direction of the force applied on the head mount by each SMA-based pennate unit. (b) The direction of the fiber force in each SMA fiber (spring) and the angle of pennation ($\alpha$) subtends with each tendon.

Meanwhile, the $F_k$ is modeled using the General stiffness model [16,17], where the ratio of a differential change in force $F_k$ at the point of termination to a differential change in the point, the position is considered the pennate stiffness ($k_t$).

$$k_t = \frac{dF_k}{dx} \quad (13)$$

$$F_k = k_t \cdot \Delta x \quad (14)$$

*6) Backbone Modelling*
Assuming the effect of shear to be negligible, we applied the Euler-Bernoulli beam theory to model the continuum backbone by considering only bending and twisting motions [18]. Correspondingly, the moment due to the backbone elasticity has been found out:

$$m = R_z(\phi) \cdot R_y(\kappa l) \begin{bmatrix} EI_{xx} & 0 & 0 \\ 0 & EI_{yy} & 0 \\ 0 & 0 & GJ/l \end{bmatrix} \begin{bmatrix} 0 \\ \kappa \\ \epsilon \end{bmatrix} \quad (15)$$

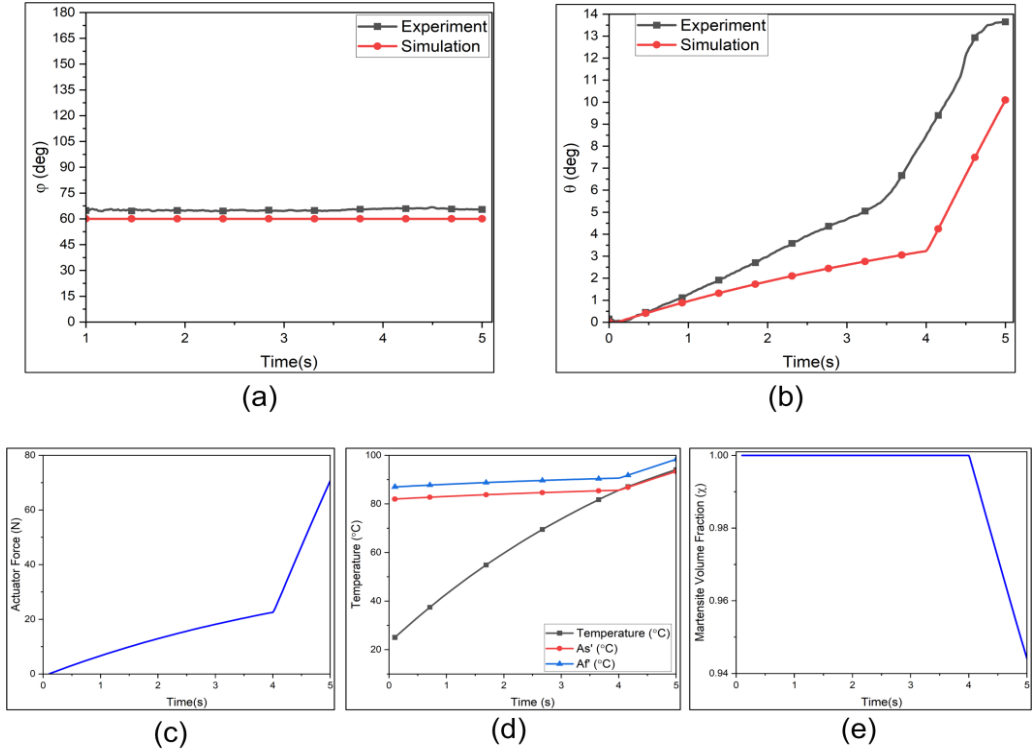

Fig 5: The comparison plot depicts the bending angle ($\theta$) and the azimuth angle ($\phi$) for both experiment (black) and simulation case (red). In this instance, 5 A has been provided for 5 seconds to one of the pennate structures. (a) The graph depicts the azimuth angle ($\phi$), as obtained via experiments and via simulation (b) The graph depicts the bending angle ($\theta$) ), as obtained via experiments and via simulation. (c) The force applied at the point of termination of each pennate structure. (d) The graph depicts the SMA spring temperature along with stress altered Austenite start ($A_s'$) and finish ($A_f'$) temperature. The martensite volume fraction starts decreasing once the wire temperature crosses the $A_s'$ as illustrated in (e).

## C. Electrical Architecture

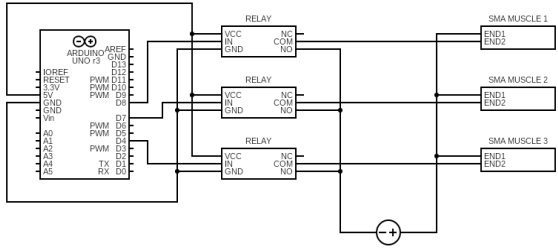

Fig. 6. Electrical Circuit, depicting the control of the three sets of SMA muscles using an Arduino UNO microcontroller via three independent relays.

The three sets of SMA muscles were independently controlled using Arduino UNO boards via three Single Pole - Double Throw (SPDT) sealed 5V Sugar Cube relays. A programmable DC power supply was used to provide required current inputs and actuate the SMA coils. The electrical architecture for actuation is depicted in Fig. 6.

The instantaneous orientation of head was measured using the MPU6050 inertial measurement sensor. The acceleration and gyroscopic measurements, retrieved from the sensor, was used to compute the tilt angles of the head mount.

## III. RESULTS

The prototype depicted in Fig 1(b), has been experimentally investigated to assess the performance of the multipennate SMA-based anthropomorphic neck. The analytical model was simulated on MATLAB and SIMULINK, and was further validated against the experimental findings, thus, establishing the validity of the analytical model. This allows us to accurately predict and control the movement of the neck. Furthermore, to demonstrate the full range of motion described by the neck, the behavior of the system under different input signals were recorded.

### A. Validation of the SMA analytical model

Fig. 5 a and 5 b presents the validation of the simulation results against the experimental results, for the input condition of 5A current input at room temperature. The $\phi$ value is an indication of which of the three SMA muscle is actuated, and the $\theta$ value expresses the degree of actuation. The $\phi$ value initially returns noise but soon stabilizes and remains nearly constant, exactly capturing the active SMA muscle. The small offset of the experimental from the simulation output, resulted from misalignment of the IMU sensor. The $\theta$ value also shows a similar trend in experimental and simulation results, with a minor deviation.

The SMA spring model accurately exhibits the SMA behavior as obtained from the analytical relationships. In Fig. 5 c, we can observe that the spring undergoes heating and as soon as condition 4 is satisfied at t = 3.99 seconds, the martensite volume fraction starts to fall (Fig. 5 d), resulting in actuation of the SMA spring. It is because of this we observe an increased muscle force and an accelerated actuation after 3.99 seconds in Fig.s 5 e and 5 b respectively.

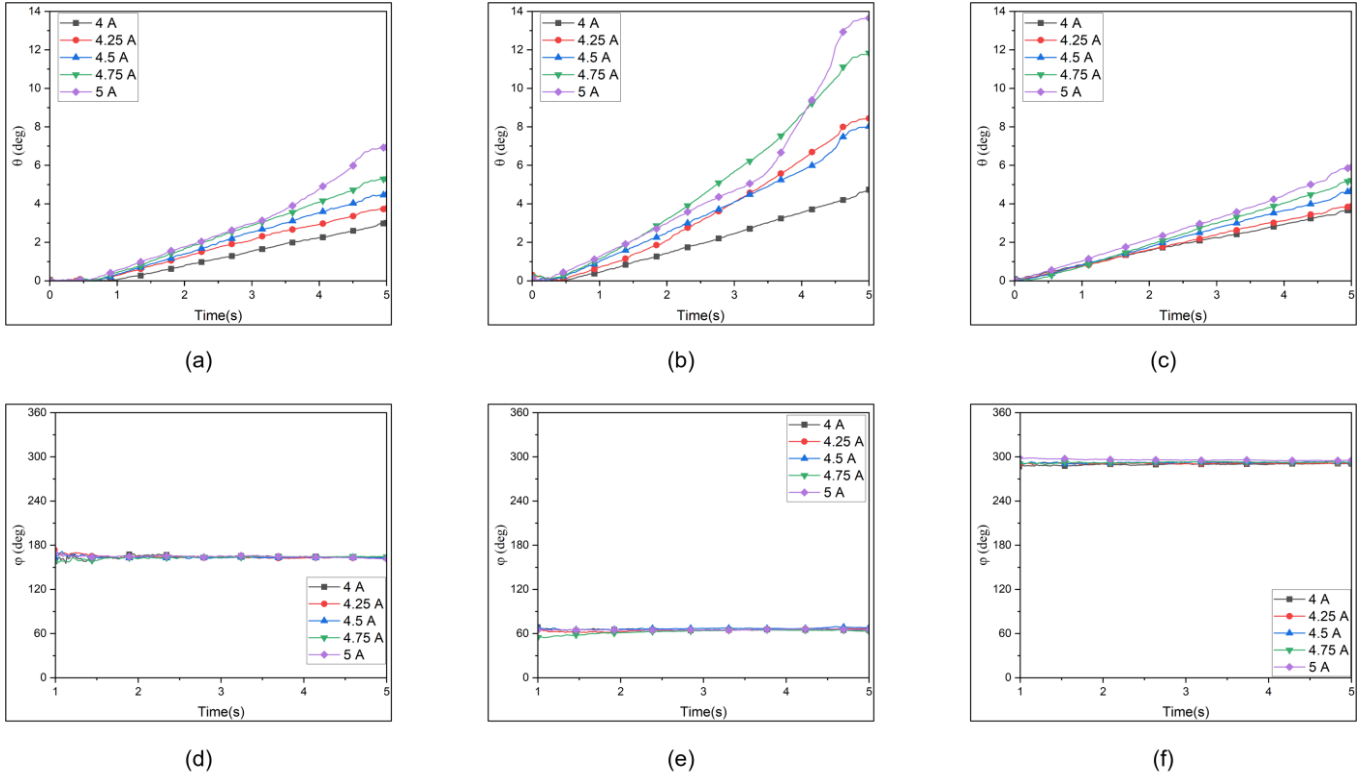

Fig 7: The graphs depict the bending angle ($\theta$) and the azimuth angle ($\phi$) for multiple instances, starting with 4 A and incremented by 0.25A, ending at 5 A. It is imperative to note that the azimuth angle ($\phi$) has been visualized after 1 second, since the azimuth angle stabilizes after the mentioned time.

Thus, a comparison between the simulation results and experiment helps us establish how well the model aligns with the developed system.

*B. Experimental results*

Two sets of experiments were conducted to observe the effect of different input current signals on the three SMA muscles and thereby demonstrating the full range of motion that the system is capable of exhibiting. In the first set, the five different input signals were chosen corresponding to 4 A, 4.25 A, 4.5 A, 4.75 A and 5 A. The resultant $\phi$ and $\theta$ values are presented in Fig. 7. A clear trend of increasing $\theta$ with increase in current can be observed. The $\phi$, on the other hand, initially shows noise and then soon stabilizes and remains constant for a particular actuated SMA muscle. Ideally, the three values at which $\phi$ stabilizes, should have been $180^o, 60^o$ and $300^o$, indicating the positions of the three SMA muscles. However, the initial orientation of the IMU sensor not being axis-aligned induces an inherent offset that can be mitigated using adaptive feedback control strategy in further development. This steady and gradual output ensures ease in precisely controlling the movement of the neck.

The next set of experiments were conducted to observe the work volume of the system. Input currents as high as 8 A can be given to obtain large bending angles without any jerk.

Table 1 reports the maximum $\theta$ values obtained for different current inputs. Comparing this to the workspace of a human neck (Fig. 1.), we can attain the full range of motion for an anthropomorphic neck.

| S No. | Input Current (A) | Maximum Bending Angle (º) |
|---|---|---|
| 1 | 4 | 4.73 º |
| 2 | 5 | 5.89 º |
| 3 | 6 | 11.34 º |
| 4 | 7 | 24.92 º |
| 5 | 8 | 32.41 º |

Table 1: The table depicts the maximum bending angle with the input current. The maximum bending angle, as predicted, increases with the input current. It is imperative to note that for each case the current was passed through the actuator for 5 seconds.

IV. DISCUSSION

This paper presents an innovative way to model the anthropomorphic neck using SMA spring arranged in the form of fibers present in multipennate muscle tissue. The pennate muscle architecture has a distinct advantage over the parallel compared to the latter in a given cross section area. The SMA muscle architecture as it generates a considerably higher force springs are arranged in a multipennate architecture, which can be imagined as a set of three muscles structures, positioned at 120° from each other. Through this configuration, a 2 DOF neck has been developed where different bending angle and azimuth angle can be achieved. Additionally, a mathematical model was developed for the SMA-based multipennate neck in the SIMULINK environment and the force generated by each pennate unit was transferred to the continuum neck model. The continuum backbone was modelled as Euler Bernoulli beam, and was approximated as an arc with constant curvature. The analytical modelling was followed by the development of the prototype. The neck prototype was fabricated with the help of additive manufacturing with each component being built with different components. It is particularly important to note that the backbone, fabricated using Thermoplastic Polyurethane (TPU) material, was given a serrated structure to provide low flexural rigidity and high axial rigidity. The analytical models were further validated by experiments and the experimental data (bending angle and azimuth angle). These were found to be consistent with the simulation results, thus validating the analytical model.

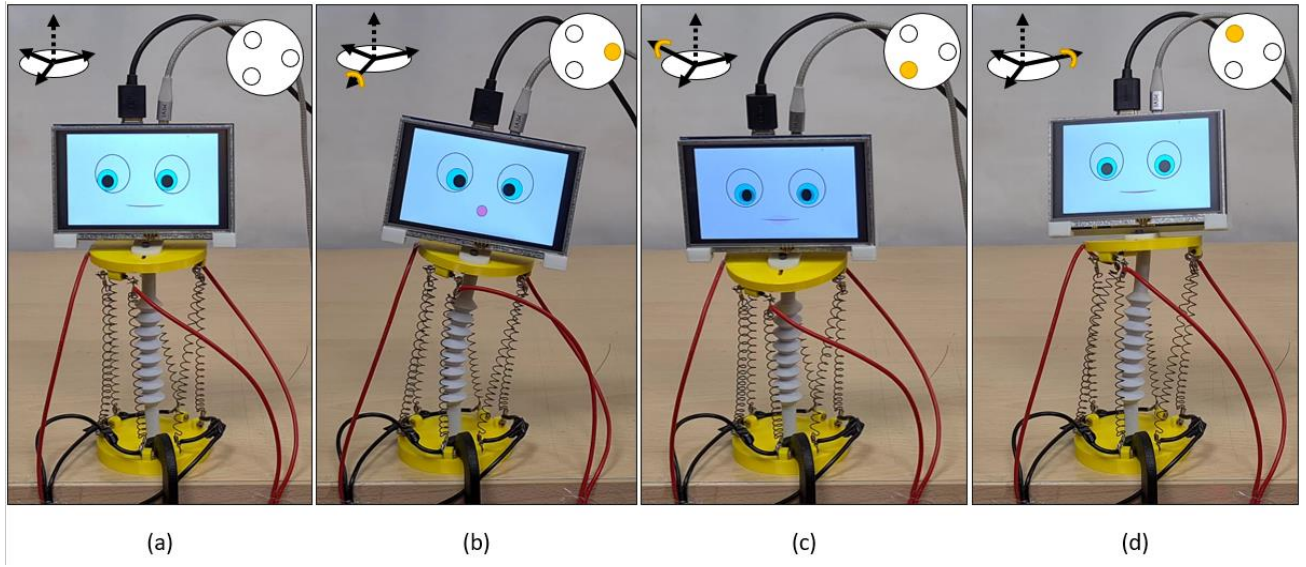

Fig. 8: The Fig. depicts the working of the bioinspired multipennate based anthropomorphic neck. (a) The image illustrate a stationary (no-actuation) stage. (b),(c),(d) depict stages of actuation, where in each stage different pennate structure has been actuated. The top right circle in each Fig. denotes the pennate unit being actuated, where yellow denotes actuation and the top left axes denote the axis of rotation for each set of actuations.

As illustrated in Fig. 5 (c), (d), the force and the stress altered SMA transition temperatures increased drastically once the martensite volume fraction starts decreasing. A simulation force of 70 N was observed via simulation and the temperature of the wire reached to 90 ℃. Once the temperature of the wire crosses the stress altered Austenite start ($A_s'$) at 3.99 seconds, the martensite volume fraction gradually starts decreasing and falls to 0.94 after 5 seconds. Moreover, experiments were conducted, with current input varying from 4 A to 5 A, where the increment was of 0.25 A. The azimuth angle ($\phi$), as expected, had three distinct values of 60°, 180° and 300° but with slight deviation due to the misalignment of the IMU sensor. Also, a maximum bending angle of 13.9° was achieved for a current of 5 A for 5 seconds. It is imperative to note that the bending angle increased with increase in input current which was also evident from Table 1, the bending angle increased with increasing the input current, since higher input current implied higher level of actuation. A maximum of 32.41° was observed for a current input of 8 A applied for 5 seconds. As illustrated in Fig 1, the anthropomorphic neck allows movements like flexion, extension and lateral flexion with a range up to 30°, which is achieved by the proposed neck in a power efficient mode through SMA springs. Furthermore, as illustrated in Fig 77, the neck had the weight of the head mount, and head display, with the combined weight of 250 gm.

## V. Conclusion

In this paper, we have presented a novel bio-inspired SMA-actuated anthropomorphic humanoid neck. Replacing traditional servo motors, we propose SMA springs, which allow us to build a lightweight system, displaying a large power-to-weight ratio. Taking inspiration from the multipennate muscles in the human body, the SMA springs are arranged in novel multipennate muscle architecture. The muscle architecture then actuates a central flexible structure, similar to muscles actuating the cervical spine in the human neck.

The paper further provides the detailed analytical modelling of the SMA springs, the multipennate muscle architecture and the continuum neck. A validation study has been conducted to compare the analytical model against the real system. Further experiments have helped in imagining the entire range of motion of the proposed system. The system is able to replicate the tilt and roll movements of the human neck, achieving a large work volume comparable to humans.

The alignment of the simulation results with the experimental results proves that the system behaves consistently, and we can predict the output for a given input. Now, in order to obtain a desired set of outputs, future work will delve into developing the inverse kinematics model and an active control system. It will enable the proposed humanoid neck to achieve complex yet precise motions. In future we wish to move further into neck design, bringing it closer and even beyond actual humans.

## Acknowledgment

The authors would like to thank Kanhaiya Lal Chaurasiya and Abhishek Kumar Singh for providing insight and expertise that greatly assisted the research.